%% file: template.tex
\documentclass{article}

\usepackage{arxiv}
\usepackage[utf8]{inputenc} 
\usepackage[T1]{fontenc}    
\usepackage{hyperref}       
\usepackage{multirow}
\usepackage{url}            
\usepackage{booktabs}       
\usepackage{amsfonts}       
\usepackage{nicefrac}       
\usepackage{microtype}      
\usepackage{lipsum}
\usepackage{bbm, bm, amssymb}
\usepackage{graphicx}
\usepackage[accsupp]{axessibility}  
\usepackage{array}
\usepackage{amsthm}
\usepackage{thmtools, thm-restate}
\usepackage{xcolor}

\usepackage{pgfplots}
\usepgfplotslibrary{patchplots}
\pgfplotsset{compat=1.3}
\usepackage{tikz}
\newcolumntype{C}[1]{>{\centering\let\newline\\\arraybackslash\hspace{0pt}}m{#1}}
\graphicspath{ {./images/} }

\newtheorem{remark}{Remark}[section]

\title{Improving Quantization with Post-Training Model Expansion}

\author{
 Giuseppe Franco,~ 
 Pablo Monteagudo-Lago,~ 
 Ian Colbert,~ 
 Nicholas Fraser,~
 Michaela Blott
 \\
 AMD
}

\begin{document}

\maketitle

\vspace{-0.3cm}
\begin{abstract}
The size of a model has been a strong predictor of its quality, as well as its cost. As such, the trade-off between model cost and quality has been well-studied.
Post-training optimizations like quantization and pruning have typically focused on reducing the overall volume of pre-trained models to reduce inference costs while maintaining model quality. However, recent advancements have introduced optimization techniques that, interestingly, expand models post-training, increasing model size to improve quality when reducing volume. For instance, to enable 4-bit weight and activation quantization, incoherence processing often necessitates inserting online Hadamard rotations in the compute graph, and preserving highly sensitive weights often calls for additional higher precision computations. However, if application requirements cannot be met, the prevailing solution is to relax quantization constraints.
In contrast, we demonstrate post-training model expansion is a viable strategy to improve model quality within a quantization co-design space, and provide theoretical justification.
In particular, when quantizing the weights and activations to $4$ bits for Llama3 1B, we reduce the gap to full-precision perplexity by an average of $9\%$ relative to both QuaRot and SpinQuant with only $5\%$ more parameters, which is still a $3.8\times$ reduction in volume relative to a BF16 reference model.
\end{abstract}

\vspace{-0.3cm}
\section{Introduction}

Quantization plays a critical role in the deployment of large language models (LLMs), offering a means of bridging the gap between full-precision reference models and their low-precision counterparts. The primary objective of quantization is to minimize inference costs while maintaining model quality by reducing the bit width requirements of weights and activations. To this end, recent advancements in post-training quantization (PTQ) have enabled the practical use of $4$-bit weights and activations during inference, effectively addressing challenges that are particularly pronounced in LLMs, such as mitigating the impact of outliers~\cite{ashkboos2024quarot, liu2024spinquant} and managing non-uniform error sensitivity~\cite{zhao2019improving,yu2024super}.

Traditionally, the quantization design space has been navigated with a fixed set of strategies; namely, the selection of bit width and data format~\cite{van2020bayesian,peters2023qbitopt,van2023fp8}, the addition or removal of zero-points~\cite{zhang2022learning,gholami2022survey}, and the granularity or optimization of scaling factors~\cite{esser2019learned, rouhani2023microscaling}.
Interestingly, parameter volume (\textit{i.e.}, model size $\times$ bit width) is commonly viewed in one dimension; volume is primarily reduced via bit width reductions, and model size is rarely considered outside of scaling analyses~\cite{frantar2022gptq, colbert2024accumulator, liu2025paretoq}.
However, in real-world applications where maintaining model quality is paramount, a modest model size increase of $5$-$10\%$ is often an acceptable trade-off.
For example, when deploying on hardware accelerators restricted to power-of-two bit widths (\textit{e.g.}, GPUs), failing to meet an accuracy requirement at $4$-bit necessitates reverting to $8$-bit precision, a significant step size increase in parameter volume.
While increasing model size typically requires either flexible supernetworks~\cite{cai2019once} or end-to-end re-training~\cite{liu2025paretoq}, we present post-training model expansion as an emerging strategy to reduce the gap between full-precision reference models and their quantized counterparts.

Our proposal aligns with existing literature,
which consistently demonstrates that increasing model size is generally an effective strategy for improving accuracy~\cite{hestness2017deep, rosenfeld2019constructive, kaplan2020scaling}. 
Moreover, a new trend is emerging; as further discussed in Section~\ref{sec:related},
recent studies have already introduced post-training quantization techniques that incidentally increase model size while still decreasing parameter volume.
We argue that post-training model expansion offers a promising path to balancing the trade-off between model cost and quality.
We propose a simple method to control  model size expansion rates via online Hadamard transformations with theoretical justification (see Section~\ref{sec:headings}),
and show that it can further improve the latest state-of-the-art weight-activation quantization algorithms,
namely,  QuaRot~\cite{ashkboos2024quarot} and SpinQuant~\cite{liu2024spinquant} (see Section~\ref{exp_setup}).
To facilitate further research and application, we have open-sourced our implementations in the Brevitas quantization library~\cite{brevitas},
complete with detailed instructions for reproduction.

\section{Related Work}
\label{sec:related}

GPTQ (also known as OPTQ)~\cite{frantar2022gptq} is the most popular and often the most effective PTQ algorithm for LLM quantization, especially when considering the \texttt{act-order} trick~\cite{brevitas, zhang2025provable}, where columns are quantized in descending order of the diagonal of the Hessian.
Recent techniques, such as QuaRot~\cite{ashkboos2024quarot}, FrameQuant~\cite{adepu2024framequant}, and QuIP~\cite{chee2024quip, tseng2024quip}, heavily rely on incoherence processing, where orthogonal rotations are introduced to reduce outliers in both weights and activations.
Unlike GPTQ, which alters weight values through iterative error correction, incoherence processing comes at the cost of adding online matrix multiplications to the compute graph (usually Hadamard matrices).
Liu \textit{et al.}~\cite{liu2024spinquant} extend incoherence processing to gradient-based PTQ, showing one can optimize rotation matrices via Cayley transforms~\cite{li2020efficient}, which preserves the orthogonality of merged rotations at no extra inference cost compared to QuaRot but with additional calibration costs.
Both QuaRot and SpinQuant start from the assumption that LLM quantization is hindered by the presence of outliers, a phenomenon studied in prior works~\cite{yu2024super}. In particular, Yu \textit{et al.}~\cite{yu2024super} empirically identify  weights that are extremely sensitive to quantization, and propose to keep them in higher precision; interestingly, these outliers can be linked to the activations' outliers.
Preserving high-precision weights limits the impact of quantization on model accuracy at the cost of high-precision matrix multiplications.
Finally, Zhao \textit{et al.}~\cite{zhao2019improving}, although not applied in the LLM context, showed that it is possible to split the most difficult channels to quantize in a layer, thus removing potential outliers and making quantization easier at the cost of slightly larger matrix multiplications.
Each of these works incidentally increase model size to improve model quality while reducing model volume.
We refer to this trend as post-training model expansion and propose a simple mechanism inspired by the intersection of these works.

\vspace{-0.3cm}
\section{Post-Training Model Expansion via Online Incoherence Processing}
\label{sec:headings}
\vspace{-0.3cm}

Incoherence processing is known to reduce the negative impact of outliers when quantizing weights and activations to low-precision datatypes, even when using per-channel and per-token scaling factor granularities, respectively~\cite{ashkboos2024quarot, liu2024spinquant}.
Given orthogonal rotation matrix $\bm{R}$, these techniques exploit the following rotation invariance property in linear layers
$$
\bm{X} \bm{W} = (\bm{XR}) (\bm{R^TW}) = \bm{XR} \bm{R}^{-1}\bm{W} = \bm{X}\bm{W}~,
$$
where, for inputs $\bm{X}$ and weights $\bm{W}$, $\bm{R}$ effectively rotates the latent space of an LLM before quantization, amortizing outliers across the rotated space before quantizing $\bm{XR}$ and $\bm{R}^T\bm{W}$.
Note that, for all orthogonal matrices, its inverse can be computed through transposition (\textit{i.e}., \( \bm{R}^T = \bm{R}^{-1} \)).
Furthermore, many rotation matrices can be merged into the linear layers around the activation quantizer, as exploited by~\cite{ashkboos2024quarot} and~\cite{liu2024spinquant}; however, this is not always possible because of non-linearities in the network.
Thus, in some circumstances, the rotation matrix is unavoidably left in the compute graph and $\mathcal{Q}(\bm{XR}) \mathcal{Q}(\bm{\tilde{W}})$ is computed during inference, where $\mathcal{Q}$ is a known quantization function and $\bm{\tilde{W}} = \bm{R}^T\bm{W}$.
We refer to this as \textit{online incoherence processing}.

We propose to extend the work of~\cite{ashkboos2024quarot} and~\cite{liu2024spinquant}, where instead of 
generic orthogonal matrices, Hadamard matrices $\bm{{H}}$ are used.
When compared to standard orthogonal matrix multiplications, Hadamard matrix multiplications can be orders of magnitude faster, owing to their recursive structure and bipolar values (all values are either $\{-\gamma, \gamma\}, \textrm{for } \gamma \in \mathbb{R}$).
When used as rotations to exploit this invariance, random Hadamard matrices are known to yield more consistent improvements in compression quality when compared to rotations with random orthogonal matrices, as shown in~\cite{liu2024spinquant}.
These benefits have led to the rapid adoption of Hadamard matrices in LLM quantization and motivated the development of optimized libraries such as Hadacore~\cite{hadacore}.
Our proposal it to expand the Hadamard matrices used during online incoherence processing, incidentally increasing model size post-training.
Let $\bm{X}$ be an ${D \times N}$ matrix of $D$ input samples of $N$ dimensions. Similarly, let the $N \times N'$ weight matrix $\bm{W}$ have $N$ input channels and $N'$ output channels.
Our proposal is to first generate an expanded $M \times M$ Hadamard matrix $\bm{H}$ where $M > N$, and then select only the first $N$ rows, yielding an ${N \times M}$ matrix $\bm{\hat{H}}$, where $\gamma=1/\sqrt{M}$. The left inverse of $\bm{\hat{H}}$ can still be computed through transposition since $\bm{\hat{H}} \bm{\hat{H}^T}= I$, allowing for fast matrix computations.
By combining $\bm{\hat{H}}$ with the weights, we obtain $\bm{\hat{W}}= \bm{\hat{H}}^T\bm{W}$ as an ${M \times N'}$ matrix, effectively increasing the number of input channels from $N$ to $M$; similarly, after the expanded Hadamard multiplication, the expanded input activation $\bm{\hat{X}} = \bm{X\hat{H}}$ is a ${D \times M}$ matrix.
Since $\bm{\hat{W}}$ and $\bm{\hat{X}}$ are linear combinations of the original $\bm{W}$ and $\bm{X}$, respectively, this also effectively increases their bit rate.
As such, the placement of these expanded Hadamard matrices is crucial to avoid an excessive increase in computational complexity without substantial benefit in model quality, as further discussed in Section \ref{exp_setup}.

\subsection{Theoretical Remarks}
\label{sec:remarks}

We present two theoretical remarks that provide insights into the nature of our post-training model expansion proposal: (1) we show that post-training model expansion strictly increases the nullspace of the input activations, which presents opportunities for carefully designed quantizers to exploit; (2) we then demonstrate that expansion indeed reduces the smallest upper bound on GPTQ reconstruction error (\textit{i.e.}, its supremum). Although our proposal focuses on Hadamard matrices, our theoretical analysis holds for any orthonormal rotation matrix.

\begin{remark}[Expansion Increases the Nullspace of the Input Activations]
\label{remark:nullspace}
\textnormal{
Recall our expanded rotation matrix $\bm{\hat{H}} \in \mathbb{R}^{N \times M}$, where $M > N$.
For input activations $\bm{X} \in \mathbb{R}^{D \times N}$, the expanded rotation matrix yields expanded input activations $\bm{\hat{X}} = \bm{X\hat{H}} \in \mathbb{R}^{D \times M}$.
By the rank-nullity theorem, it follows that
$$
N = \textrm{rank}(\bm{X}) + \textrm{nullity}(\bm{X}) , \quad
M = \textrm{rank}(\bm{\hat{X}}) + \textrm{nullity}(\bm{\hat{X}}).
$$
Since $\bm{\hat{H}}$ is full-row rank by construction, it then follows that $\textrm{rank}(\bm{\hat{X}}) = \textrm{rank}(\bm{{X\hat{H}}}) = \textrm{rank}(\bm{{X}})$, and therefore
$$
\textrm{nullity}(\bm{\hat{X}}) - \textrm{nullity}(\bm{X}) = M - N > 0.
$$
Thus, there is a strictly positive increase in the dimensionality of the nullspace of the input activations, namely
$$
\textrm{nullity}(\bm{\hat{X}}) \geq \textrm{nullity}(\bm{X}).
$$
Interestingly, this implies that post-training model expansion incidentally grants more opportunity to effectively hide quantization error during calibration.
As our goal is to find a quantization mapping $\mathcal{Q}$ such that $\bm{X} \bm{W} = \bm{X} \mathcal{Q}(\bm{W})$ or, equivalently, that $\bm{w}-\mathcal{Q}(\bm{w}) \in  \text{nullspace}( \bm{X} )$ (for every column $\bm{w}$ in $\bm{W}$), one could expect increasing the nullspace of the input feature space to help reduce the quantization error. In other words, though rotation does not alter the intrinsic dimensionality of the data, the higher redundancy in the projected space could be exploited by a carefully designed quantizer, which we further discuss in Remark \ref{remark:error} for GPTQ.}
\end{remark}

\begin{remark}[Expansion Can Decrease GPTQ Reconstruction Error]
\label{remark:error}
\textnormal{It is well known that rotations do not alter the column space of a matrix, nor its energy. More formally, given input data $\bm{X} \in \mathbb{R}^{D \times N}$ and orthonormal matrix~\( \bm{H} \in \mathbb{R}^{N \times N} \), one can show that \( P_{\bm{XH}^\perp} = P_{\bm{X}^\perp} \) and \( \Vert \bm{XH} \Vert_F^2 = \Vert \bm{X} \Vert_F^2 \), where \( P_{\bm{X}^\perp} \) is the projection onto the orthogonal complement of the column space of $\bm{X}$. Thus, given that $\bm{\tilde{w}} \in \mathbb{R}^N$ is a column from weight matrix \( \bm{\tilde{W}} = \bm{H}^T\bm{W} \) and \( \bm{q} \) is its quantized counterpart, it follows that GPTQ yields the following error bound via Proposition 3.2 of \cite{zhang2025provable}
\begin{equation}
\Vert \bm{XH\tilde{w}} - \bm{XHq} \Vert_2 \leq \Delta \sqrt{N} \cdot \min \left\{ \max_j \Vert P_{\bm{X}_{\geq j+1}^\perp} \bm{XH}_j \Vert_2 , \sqrt{\frac{\Vert \bm{X} \Vert_F^2}{N}}  \right\},
\end{equation}
where \( \bm{X} \) is assumed to be full-column rank with \( D \geq N \), \( \bm{H}_j \) is the \( j^{th} \) column of \( \bm{H} \) and \( \Delta \) is the maximum rounding error for known quantization function~\( \mathcal{Q} \).
Similar to \( \bm{H} \), one can show that our expanded rotation matrix \( \hat{\bm{H}} \in \mathbb{R}^{N \times M} \) also leaves the column space and energy of a matrix unaltered such that \( \Vert \bm{X\hat{H}} \Vert_F^2 = \Vert \bm{X} \Vert_F^2 \) and \( P_{\bm{X\hat{H}}^\perp} = P_{\bm{X}^\perp} \),  while also preserving its rank (see Remark~\ref{remark:nullspace}). Thus, given that \( \hat{\bm{w}} \in \mathbb{R}^{M} \) is a column from \( \hat{\bm{W}} = \hat{\bm{H}}^T \bm{W} \), it also follows that GPTQ yields the following error bound with expanded rotations
\begin{equation}
\Vert \bm{X\hat{H}\hat{w}} - \bm{X\hat{H}q} \Vert_2 \leq \Delta \sqrt{N} \cdot \min \left\{ \max_j \Vert P_{\bm{X}_{\geq j+1}^\perp} \bm{X\hat{H}}_j \Vert_2 , \sqrt{\frac{\Vert \bm{X} \Vert_F^2}{N}}  \right\}.
\end{equation}
Noticeably, the worst-case error is unperturbed. However, since \( \Vert \bm{H}_j \Vert_2 =1 \) and \( \Vert \bm{\hat{H}}_j \Vert_2 = \sqrt{\frac{N}{M}} \) for all \( j \), one can show via the spectral norm bound that
\[ \max_j \Vert P_{\bm{X}_{\geq j+1}^\perp} \bm{XH}_j \Vert_2  \leq \max_j \Vert P_{\bm{X}_{\geq j+1}^\perp} \bm{X} \Vert_2 \Vert \bm{H}_j \Vert_2 = \max_j  \Vert P_{\bm{X}_{\geq j+1}^\perp} \bm{X} \Vert_2   \]
and
\[ \max_j \Vert P_{\bm{X}_{\geq j+1}^\perp} \bm{X\hat{H}}_j \Vert_2  \leq \max_j  \Vert P_{\bm{X}_{\geq j+1}^\perp} \bm{X} \Vert_2 \Vert \bm{\hat{H}}_j \Vert_2 = \sqrt{\frac{N}{M}} \max_j \Vert P_{\bm{X}_{\geq j+1}^\perp} \bm{X} \Vert_2.   \]
Thus, our analysis admits a reduction of the supremum of GPTQ reconstruction error by a factor of \( \textstyle \sqrt{N / M} \) via post-training model expansion.
Although this does not guarantee universal improvement at this rate, our analysis does establish a theoretical rate of decay for the worst-case error under ideal conditions. While we offer no theoretical insights into how these ideal conditions may manifest in practical settings, we do provide empirical evidence in Section~\ref{results} that shows our post-training model expansion technique indeed reduces quantization error.}
\end{remark}

\section{Experimental Setup}
\label{exp_setup}

\textbf{Models \& Datasets.} We evaluate our post-training model expansion technique on the Llama 3 family of models~\cite{grattafiori2024llama}, focusing in particular on Llama 3.2 1B and 3B, and Llama 3.1 8B. 

We use WikiText2~\cite{merity2016pointer} as the dataset for calibration and evaluation with 128 training samples used for calibration and a sequence length of 2048 tokens, a common configuration used in prior works~\cite{liu2024spinquant, frantar2022gptq, colbert2024accumulator}.
For zero-shot evaluation, we considered the following tasks: ARC-easy (ARC-E) and ARC-challenge (ARC-C)~\cite{clark2018think}, HellaSwag (HS)~\cite{zellers2019hellaswag}, PIQA~\cite{bisk2020piqa}, and Winogrande (Wino)~\cite{sakaguchi2021winogrande}.
 We report the unnormalized results for all tests using Huggingface's LightEval library~\cite{lighteval}.

\textbf{Quantization Details.} We consider two quantization configurations.
First, we consider the 4-bit weight and activation quantization paradigm proposed by Liu \textit{et. al}~\cite{liu2024spinquant}, which uses per-channel symmetric weight quantization and per-token asymmetric activation quantization.
Second, we consider MXFP4 for both weights and activations as specified by the Open Compute Project (OCP), which applies E8M0 microscaling (MX) for each group of 32 elements \cite{ocp2024mx}.
We apply GPTQ~\cite{frantar2022gptq} for weight quantization and quantize weight columns in descending order of the diagonal of the Hessian (\textit{i.e.}, the \texttt{act-order} trick~\cite{brevitas}).
We do not quantize the KV cache as expansion does not directly impact it.

\textbf{Compute Graph Details.}
We evaluate two different compute graph modifications. First, we consider the same structure proposed by Liu \textit{et. al}~\cite{liu2024spinquant}.
In this case, rotation matrices are merged into linear layers when possible and only the down projection layer requires online incoherence processing (see Section~\ref{sec:headings}).
The main difference from~\cite{liu2024spinquant} lies in the expansion of the down projection layer ($R_4$ by the notation in~\cite{liu2024spinquant}), which does not impact the placement or number of standalone Hadamard rotations.
Second, we experiment with expanding the output projection layer rotation in the scaled dot product attention (SDPA) module ($R_2$ by the notation in~\cite{liu2024spinquant}), which localizes the impact of expansion.

\textbf{Hadamard Optimization.}
We perform two main sets of experiments, denoted as QuaRot* and SpinQuant*, which respectively re-implement the proposals from \cite{ashkboos2024quarot} and \cite{liu2024spinquant}.
When evaluating Cayley optimization in SpinQuant*, we used 800 training samples and the same training hyperparameters reported in \cite{liu2024spinquant}.
When using QuaRot*, we use a deterministic Hadamard matrix rather than using random Hadamard matrices.

All algorithms are implemented in Brevitas~\cite{brevitas}, and we provide instructions on how to replicate our results.
For each of our experiments, we report the model sizes for the quantized variants and the ratio of expansion compared to the base quantized version. These are computed by counting the total number of parameters in the network. We exclude the last linear layer from this count since its parameters are shared with the embedding layers. Similarly, we report the number of input channels for the down projection layers, and their expansion ratios.
The ratio 1.0 refers to the baseline quantization configuration (\textit{i.e.}, QuaRot* or SpinQuant*), with no extra online Hadamard expansion.

\begin{table}[h]
\centering
\caption{\textbf{Expanding $R_4$ with QuaRot* for Llama3 models}. We compare different post-training expansion rates for perplexity (lower is better) and several zero-shot tasks (higher is better). We also report the overall model size of the different quantized configurations and the expansion ratio compared to baseline quantized. Similarly, we report the input channels for the down projection layers and the expansion ratio.}
\begin{tabular}{ccC{2cm}C{2cm}cccccc}
\toprule
Model                      & Algorithm & Model Size (Ratio)                  & Layer size (Ratio) & Perplexity &  ARC-C  & ARC-E &  HS   & Wino  & PIQA        \\ \midrule
\multirow{6}{*}{1B} & Float                          & -         & -                       &   8.94   &  31.91  & 66.45  & 48.15  & 60.06  &  75.46  \\  \cmidrule(l){2-10}
                           & \multirow{5}{*}{QuaRot*}       &  1.235B (1.00)  & 8192  (1.00)         & 12.94   &  26.02  & 56.14 & 41.39 & 54.70 & 67.57   \\ 
                           &                                &  1.260B (1.02)  & 8960  (1.10)         & 12.75   &  28.75  & 57.82 & 41.85 & 55.01 & 68.39   \\ 
                           &                                &  1.285B (1.04)  & 9728  (1.20)         & 12.75   &  29.35  & 58.79 & 41.76 & 55.17 & 69.47     \\ 
                           &                                &  1.302B (1.05)  & 10240 (1.30)         & 12.56   &  26.88  & 55.89 & 40.82 & 53.51 & 67.95     \\ 
                           &                                &  1.369B (1.11)  & 12288 (1.50)         & 12.56   &  27.64  & 56.69 & 41.17 & 56.35 & 69.10     \\ \midrule
\multirow{6}{*}{3B} & Float                          & -                        & - &   7.16   & 42.40   &  74.87 &  55.83  & 68.43  & 76.70    \\ \cmidrule(l){2-10}
                           & \multirow{5}{*}{QuaRot*}       &  3.212B (1.00)   & 8192 (1.00)  &    9.19  &  36.86  & 67.34 & 49.96 & 59.51 & 73.07   \\ 
                           &                                &  3.279B (1.05)  & 8960 (1.02)  &    9.19   &  35.83  & 68.48 & 49.73 & 61.09 & 73.12  \\ 
                           &                                &  3.345B (1.05)  & 9728 (1.04)  &    9.19   &  36.35  & 69.28 & 49.71 & 62.12 & 73.18  \\ 
                           &                                &  3.388B (1.05)  & 10240 (1.05)  &    9.01   &  38.74  & 69.91 & 49.87 & 64.25 & 72.03  \\ 
                           &                                &  3.564B (1.08)  & 12288 (1.10) &    9.01   &  37.03  & 68.35 & 50.88 & 62.98 & 73.39     \\ \midrule
\multirow{6}{*}{8B} & Float                          & -                          & - &   5.91   &   51.79 &  82.20 &  60.73 & 70.87  & 79.11    \\ \cmidrule(l){2-10}
                           & \multirow{5}{*}{QuaRot*}       &  7.504B (1.00)   & 14336 (1.00)   &    7.50   &  44.97  & 76.80 & 54.69 & 65.11 & 75.79   \\ 
                           &                                &  7.638B (1.02)  & 15360 (1.07)  &    7.38   &  45.22  & 77.52 & 55.62 & 65.27 & 76.11     \\ 
                           &                                &  8.040B (1.07)  & 18432 (1.30)  &    7.38   &  44.54  & 77.40 & 55.06 & 65.11 & 76.27     \\ 
                           &                                &  8.243B (1.10)  & 19968 (1.40)  &    7.34   &  46.67  & 77.73 & 55.64 & 67.09 & 76.55     \\
                           &                                &  8.511B (1.13)  & 22016 (1.50)  &    7.28   &  44.11  & 77.02 & 55.27 & 66.85 & 77.04     \\ \bottomrule

\end{tabular}
\label{table:quarot}
\end{table}

\begin{table}[h]
\centering
\caption{\textbf{Expanding $R_4$ with SpinQuant* for Llama 3.2 1B.} We evaluate the impact of different expansion factors on perplexity (lower is better) and several zero-shot tasks (higher is better).}
\begin{tabular}{ccC{2cm}cccccc}
\toprule
 Model                     & Algorithm                          & Layer Size (Ratio)   & Perplexity & ARC-C       & ARC-E       & HS          & Wino        & PIQA       \\ \midrule
\multirow{5}{*}{1B} & \multirow{5}{*}{SpinQuant*}               & 8192 (1.00)      & 12.22     & 28.16       & 59.05       & 42.38       & 53.12       & 68.99     \\ 
&                                                               & 8960  (1.10)     & 12.12     & 28.07       & 58.63       & 41.78       & 54.93       & 68.50      \\ 
&                                                               & 9728  (1.20)     & 12.04     & 28.32       & 57.53       & 42.30       & 54.77       & 68.23      \\ 
&                                                               & 10240 (1.30)     & 11.93     & 28.15       & 57.15       & 42.07       & 54.22       & 67.74      \\ 
&                                                               & 12288 (1.50)     & 11.80     & 28.67       & 58.80       & 42.95       & 54.78       & 68.61      \\ \bottomrule
\end{tabular}
\label{table:spinquant}
\end{table}

\section{Results}
\label{results}

\textbf{Expanding $R_4$.}
In Table~\ref{table:quarot}, we can see that slightly expanding $R_4$, and thus slightly increasing the model size by increasing the size of the down projection layer, improves both perplexity and zero-shot performance, 
although there are quickly diminishing returns in perplexity improvements with higher expansion rates, as Remark \ref{remark:error} predicts.

In these experiments, optimization was performed using \texttt{bfloat16} as the base datatype.
Next, we show that combining Hadamard expansions with Cayley optimization for orthogonal rotation matrices can further improve these results. 
In these experiments, optimization was performed using \texttt{float32} as the base datatype, instead of \texttt{bfloat16}.
Results are provided in Table~\ref{table:spinquant}; we do not report the performance of the float model, or the model size, which are identical to Table~\ref{table:quarot}.
As expected, Cayley-optimized rotations improves on the base Hadamard rotations.
Furthermore, Cayley-optimized expanded rotations (\textit{i.e.}, SpinQuant*) consistently provides benefits over QuaRot*, as reported in~\cite{liu2024spinquant},
and narrows the gap between the quantized model and its high-precision counterpart. 
We again can see that larger expansions correlate with larger reductions in perplexity, with diminishing returns on zero-shot accuracy;
however, more experiments may be needed to draw a conclusion on whether it is possible to further improve zero-shot results.

\textbf{Expanding $R_2$ and $R_4$.}
When expanding more rotations, and therefore more layers, it is important to consider the trade-offs.
In the best case scenario, the expanded Hadamard rotation can be fused in the neighbouring layers,
and it won't have a cascading effect on the other layers of the network.
However, if that is not the case, the best solution would then be to have more standalone Hadamard rotations, which,
as mentioned before, might slow down inference speed of the network.
Furthermore, having more input channels implies larger and slower matrix multiplications for two types of layers across the entire model.
From our results, reported in Tables \ref{table:quarot_double} and \ref{table:spinquant_double},
we can see that expanding more layers provides benefits both in terms of perplexity and zero shot accuracy, also when using learned hadamard rotations.
Compared to the previous results, we only report the overall increase in model size, and not the individual
increase in layer sizes.

\begin{table}[h]
    \centering
    \caption{\textbf{Expanding $R_2$ and $R_4$ with QuaRot* for Llama3 models}. We compare different post-training expansion rates for perplexity (lower is better) and several zero-shot tasks (higher is better). We also report the overall model size of the different quantized configurations and the expansion ratio compared to baseline quantized.}
    \begin{tabular}{ccC{2cm}cccccc}
    \toprule
    Model                      & Algorithm & Model Size (Ratio)  & Perplexity &  ARC-C  & ARC-E &  HS   & Wino  & PIQA        \\ \midrule
    \multirow{3}{*}{1B} & Float                          & -                    &   8.94   &  31.91  & 66.45  & 48.15  & 60.06  &  75.46  \\  \cmidrule(l){2-9}
                               & \multirow{2}{*}{QuaRot*}       &  1.235B (1.00)  & 12.94   &  27.13  & 55.39 & 40.93 & 53.19 & 67.68   \\ 
                               &                                &  1.403B (1.14)          & 12.19   &  27.30  & 55.05 & 42.57 & 55.17 & 67.79     \\ \midrule
    \multirow{3}{*}{3B} & Float                                      & - &   7.16   & 42.40   &  74.87 &  55.83  & 68.43  & 76.70    \\ \cmidrule(l){2-9}
                               & \multirow{2}{*}{QuaRot*}       &  3.212B (1.00)   &    9.06   &  38.40  & 69.44 & 50.02 & 62.75 & 72.85   \\ 
                               &                                &  3.719B (1.16)  &    8.75    &  36.60  & 70.08 & 50.68 & 63.06 & 74.10     \\ \midrule
    \multirow{3}{*}{8B} & Float                                        & - &   5.91   &   51.79 &  82.20 &  60.73 & 70.87  & 79.11    \\ \cmidrule(l){2-9}
                               & \multirow{2}{*}{QuaRot*}       &  7.504B (1.00)    &    7.34   &  44.20  & 75.68 & 55.47 & 65.43 & 78.03   \\ 
                               &                                &  8.780B (1.17)   &    7.06   &  44.88  & 77.86 & 56.26 & 67.32 & 76.82     \\ \bottomrule
    
    \end{tabular}
    \label{table:quarot_double}
    
\end{table}
    
\begin{table}[h]
\centering
\caption{\textbf{Expanding $R_2$ and $R_4$ with SpinQuant* for Llama 3.2 1B}. We evaluate the impact of different expansion factors on perplexity (lower is better) and several zero-shot tasks (higher is better).}
\begin{tabular}{ccC{2cm}cccccc}
\toprule
    Model                     & Algorithm                          & Model Size (Ratio)   & Perplexity & ARC-C       & ARC-E       & HS          & Wino        & PIQA       \\ \midrule
\multirow{2}{*}{1B} & \multirow{2}{*}{SpinQuant*}               & 1.235B (1.00)      & 12.11     & 29.69       & 58.16       & 41.79       & 54.54       & 67.95     \\ 
&                                                               & 1.403B (1.14)     & 11.40     & 28.84       & 60.90       & 43.00       & 53.35       & 68.88      \\ \bottomrule
\end{tabular}
\label{table:spinquant_double}
\end{table}

\textbf{OCP MXFP4 Results}.
In addition to the quantization configuration proposed by Liu \textit{et. al}~\cite{liu2024spinquant}, 
we also investigate the impact of model expansion with the OCP MXFP4 datatype.
We focus our experiments on expanding both $R_2$ and $R_4$.
Compared to the quantization configuration used so far, OCP MXFP4 involves the use of more scale factors for each tensor,
for both weights and activations.
The extra computational complexity of this format is offset by the emergence of specialized hardware,
that can efficiently take advantage of the group-wise structure of the scale during the matmul operation.
This effectively negates most of the drawbacks of this datatype, while still guaranteeing higher precision compared to
quantization configurations with coarser granularity for scale factors.

\begin{table}[h]
    \centering
    \caption{\textbf{Expanding $R_2$ and $R_4$ with QuaRot* for Llama3 models quantized to OCP MXFP4}. We compare different post-training expansion rates for perplexity (lower is better) and several zero-shot tasks (higher is better). We also report the overall model size of the different quantized configurations and the expansion ratio compared to baseline quantized.}
    \begin{tabular}{ccC{2cm}cccccc}
    \toprule
    Model                      & Algorithm & Model Size (Ratio)                  & Perplexity &  ARC-C  & ARC-E &  HS   & Wino  & PIQA        \\ \midrule
    \multirow{3}{*}{1B} & Float                          & -         &   8.94  &  31.91  & 66.45  & 48.15  & 60.06  &  75.46  \\  \cmidrule(l){2-9}
                               & \multirow{2}{*}{QuaRot*}       &  1.235B (1.00)        & 11.81   &  27.05  & 59.09 & 41.37 & 56.20 & 68.82   \\ 
                               &                                &  1.403B (1.14)     & 11.44   &  28.24  & 59.93 & 42.46 & 54.06 & 70.73     \\ \midrule
    \multirow{3}{*}{3B} & Float                           & - &   7.16   & 42.40   &  74.87 &  55.83  & 68.43  & 76.70    \\ \cmidrule(l){2-9}
                               & \multirow{2}{*}{QuaRot*}       &  3.212B (1.00)     &    8.94   &  33.70  & 67.21 & 47.82 & 61.01 & 72.52   \\ 
                               &                                &  3.719B (1.16)   &    8.63    &  39.25  & 71.84 & 51.32 & 64.09 & 73.88     \\ \midrule
    \multirow{3}{*}{8B} & Float                             & - &   5.91   &   51.79 &  82.20 &  60.73 & 70.87  & 79.11    \\ \cmidrule(l){2-9}
                               & \multirow{2}{*}{QuaRot*}       &  7.504B (1.00)    &    7.10   &  44.97  & 77.81 & 56.02 & 69.06 & 76.50   \\ 
                               &                                &  8.780B (1.17)   &    6.94   &  45.90  & 78.11 & 56.64 & 70.24 & 77.04     \\ \bottomrule
    
    \end{tabular}
    \label{table:quarot_mxfp4}

\end{table}

\begin{table}[h]
\centering
\caption{\textbf{Expanding $R_2$ and $R_4$ with SpinQuant* for Llama 3.2 1B quantized to OCP MXFP4.} We evaluate the impact of different expansion factors on perplexity (lower is better) and several zero-shot tasks (higher is better).}
\begin{tabular}{ccC{2cm}cccccc}
\toprule
    Model                     & Algorithm                         & Model Size (Ratio)    & Perplexity & ARC-C       & ARC-E       & HS          & Wino        & PIQA       \\ \midrule
\multirow{2}{*}{1B} & \multirow{2}{*}{SpinQuant*}               & 1.235B (1.00)      & 11.59     & 29.52       & 59.39       & 42.23       & 55.64       & 70.95     \\ 
&                                                               &  1.403B (1.14)      & 11.34     & 30.72       & 61.07       & 42.84       & 56.99       & 70.35      \\ \bottomrule
\end{tabular}
\label{table:spinquant_mxfp4}
\end{table}

\textbf{The Trade-off Between Model Quality and Model Volume}.
Quantization often focuses on the amount of model degradation compared to some full-precision baseline.
When quantization is applied to multiple models designed for the same problem (\textit{e.g.}, multiple LLMs trained on the same data),
the effect of quantization can be placed into a wider context.
For example, if there is some constraint at inference time,
such as model volume,
we can ask: "what is the most accurate model which fits my requirements?"
Using model volume as an example constraint, Figures~\ref{fig:pareto_perplexity}~and~\ref{fig:pareto_0shot} show the trade-off that can be made between volume and WikiText2 perplexity or average zero-shot evaluation, respectively.
The figures show the results from Tables~\ref{table:quarot} and~\ref{table:spinquant},
the different clusters in the figures correspond to the different models sizes: Llama 3.2 1B, 3.2 3B, and 3.1 8B.
For a given memory constraint (measured as model volume), the Pareto frontier provides the optimal model within our co-design space.
Notably, SpinQuant results outperform QuaRot results for equivalent models (\textit{i.e.}, Llama 3.2 1B),
otherwise with respect to Figure~\ref{fig:pareto_perplexity}, almost all quantized and float models lie on the Pareto frontier.
Interestingly, for the zero-shot results, the Llama 3.2 1B and 3B \texttt{bfloat16} results are completely Pareto-dominated by the INT4 Llama 3.2 3B and 8B results respectively.
Finally, since all experiments quantize both weights and activations,
we expect the trade-off between model volume and model quality would significantly lean further towards quantized models
if weight-only quantization were considered.

\begin{figure}[h]
\centering
\input{images/pareto}
\label{fig:pareto}
\end{figure}

\section{Conclusion}

Post-training model expansion broadens the quantization landscape. 
When accuracy requirements cannot be met, the most common strategy is to relax quantization constraints,  for example going from {per-tensor} to {per-channel} (or even {per-group}) scaling or increasing bit widths.
However, compilers and accelerators are often  designed and optimized to work best with a specific set of datatypes.
Therefore, as an alternative strategy, we demonstrate modest increases in model size post-training can improve the accuracy of quantized models within a fixed quantization co-design space.
We offer theoretical insights into why and how this may happen via analysis of the expanded nullspace and its impact on GPTQ reconstruction error bounds.
Our simple post-training model expansion trick gives more flexibility to deep learning engineers deploying LLMs, and our investigations open up promising new research questions around the potential for data-dependent quantization algorithms to overfit, and the general performance plateau of data-free quantization.
Furthermore, one could precisely target specific layers based on their sensitivity to quantization; we leave such a study for future work.
Finally, although Cayley optimization seems to consistently work well with Hadamard expansion, we plan to investigate the variability in zero-shot results that we observed when testing for different model expansion rates.

\section*{Acknowledgments}

We would like to thank Shihao Zhang of UC San Diego for discussions on the GPTQ reconstruction error bounds.

\bibliographystyle{unsrt}  
\bibliography{template}

\end{document}

%% file: images/pareto.tex
\begin{minipage}{0.4\linewidth}
\hspace{-0.5cm}
\begin{tikzpicture}[scale=0.95]
    \begin{axis}[
        xmin=0,
        xmax=16,
        ymin=5,
        ymax=15,
        xlabel=Model Volume (GB),
        ylabel=Perplexity,
        grid=both,
        legend cell align=left,
        legend style={legend pos=north east,font=\small}
    ]
        \addplot[dashed,mark=none,color=black,thick] table [x=size, y=perplexity, col sep=comma] {./images/perp_pareto.csv}; \addlegendentry{Pareto Frontier};
        \addplot[only marks,mark=*,color=blue,mark size=2.5pt] table [x=size, y=perplexity, col sep=comma] {./images/perp_float.csv}; \addlegendentry{BF16};
        \addplot[only marks,mark=square*,color=red,mark size=2.5pt] table [x=size, y=perplexity, col sep=comma] {./images/perp_quarot.csv}; \addlegendentry{INT4 (QuaRot*)};
        \addplot[only marks,mark=triangle*,color=green!50!black,mark size=2.5pt] table [x=size, y=perplexity, col sep=comma] {./images/perp_spinquant.csv}; \addlegendentry{INT4 (SpinQuant*)};
    \end{axis}
\end{tikzpicture}
\caption{Comparison between Wikitext2 perplexity scores and model volume in GB.}
\label{fig:pareto_perplexity}
\end{minipage}
~
\hspace{1cm}
\begin{minipage}{0.4\linewidth}
\vspace{-0.3cm}
\begin{tikzpicture}[scale=0.95]
    \begin{axis}[
        xmin=0,
        xmax=16,
        ymin=45,
        ymax=70,
        xlabel=Model Volume (GB),
        ylabel=Avg. 0-Shot Accuracy (\%),
        grid=both,
        legend cell align=left,
        legend style={legend pos=south east,font=\small}
    ]
        \addplot[dashed,mark=none,color=black,thick] table [x=size, y=geomean, col sep=comma] {./images/0shot_pareto.csv}; \addlegendentry{Pareto Frontier};
        \addplot[only marks,mark=*,color=blue,mark size=2.5pt] table [x=size, y=geomean, col sep=comma] {./images/0shot_float.csv}; \addlegendentry{BF16};
        \addplot[only marks,mark=square*,color=red,mark size=2.5pt] table [x=size, y=geomean, col sep=comma] {./images/0shot_quarot.csv}; \addlegendentry{INT4 (QuaRot*)};
        \addplot[only marks,mark=triangle*,color=green!50!black,mark size=2.5pt] table [x=size, y=geomean, col sep=comma] {./images/0shot_spinquant.csv}; \addlegendentry{INT4 (SpinQuant*)};
    \end{axis}
\end{tikzpicture}
\caption{Comparison between geometric mean of 0-shot evaluation and model volume in GB.}
\label{fig:pareto_0shot}
\end{minipage}